\documentclass{article}
\usepackage{ijcai23}

\usepackage{times}
\usepackage{soul}
\usepackage{url}
\usepackage[hidelinks]{hyperref}
\usepackage[utf8]{inputenc}
\usepackage[small]{caption}
\usepackage{graphicx}
\usepackage{amsmath}
\usepackage{amsthm}
\usepackage{booktabs}
\usepackage{algorithm}
\usepackage{algorithmic}
\usepackage[switch]{lineno}

\newtheorem{definition}{Definition}

\title{Choose your Data Wisely: A Framework for Semantic Counterfactuals}

\author{
Edmund Dervakos$^{1}$
\and
Konstantinos Thomas$^{1}$\and
Giorgos Filandrianos$^{1}$\And
Giorgos Stamou$^1$
\affiliations
$^1$National Technical University of Athens\\
\emails
\{eddiedervakos, kthomas, geofila\}@islab.ntua.gr,
gstam@cs.ntua.gr,
}

\begin{document}

\maketitle

\begin{abstract}
    Counterfactual explanations have been argued to be one of the most intuitive forms of explanation. They are typically defined as a minimal set of edits on a given data sample that, when applied, changes the output of a model on that sample. However, a minimal set of edits is not always clear and understandable to an end-user, as it could, for instance, constitute an adversarial example (which is indistinguishable from the original data sample to an end-user). Instead, there are recent ideas that the notion of minimality in the context of counterfactuals should refer to the semantics of the data sample, and not to the feature space. In this work, we build on these ideas, and propose a framework that provides counterfactual explanations in terms of knowledge graphs. We provide an algorithm for computing such explanations (given some assumptions about the underlying knowledge), and quantitatively evaluate the framework with a user study. 
\end{abstract}

\section{Introduction}
As the field of eXplainable AI (XAI) research matures, the desiderata for explanations become more clear. Some of these are difficult to quantify, such as understandability, trust, and informativeness \cite{hoffman2018metrics}. For counterfactual explanations specifically, there are additional requirements, namely feasibility and actionability \cite{poyiadzi2020face}, minimality/proximity, and flip-rate/accuracy/validity \cite{verma2020counterfactual}. In this work, we propose a framework for generating counterfactual explanations, that guarantees optimal flip-rate, minimality, feasibility, and actionability. Furthermore, we argue that by utilizing knowledge graphs, the explanations become more understandable, informative, and trustworthy - which we quantify in a user study.

Explanations based on low-level features, such as pixel brightness or sound frequency, have not proven to be particularly helpful or trustworthy to users \cite{rudin2019stop}. These numeric representations of real-world phenomena appear rather cryptic to humans, who choose intuitive, high-level criteria when describing the factors that direct their decisions; criteria such as, how ``furry'' a dog is or how ``dry'' a cough sounds when determining a dog breed or a  respiratory issue, respectively. Low-level features may be useful information for machine predictions, but not for human-readable explanations.

Fortunately, there is no mathematical difference between a vector representing low-level characteristics (eg. pixel values) and one representing semantically rich features \cite{browne2020semantics}. This makes it feasible to create systems that provide counterfactual explanations in terms of semantic features instead of low-level characteristics. This argument has been grounded both theoretically and practically, by the community. \cite{browne2020semantics} demonstrates the equivalence between counterexamples and adversarial examples in cases where higher-level semantics are not employed. Additionally, recent works provide explanations by incorporating the semantics of inputs in various ways. For instance, \cite{goyal2019counterfactual} uses the intermediate output of a classifier as a form of higher-level information about the input image. \cite{akula2020cocox} deduces an image's semantic concepts (xconcepts) by clustering the outputs of these features and presuming that elements of the same cluster are semantically similar. Meanwhile, \cite{vandenhende2022making} employs an external neural network to create semantic embeddings of these features, regarding proximate embeddings as semantically equivalent. All of the above algorithms use different methods of approximating the semantic distance of the elements depicted in images. 

In our work, we build on these ideas and propose a framework for semantic counterfactuals, where the semantics are defined in knowledge graphs \cite{hogan2021knowledge}. This allows for providing explanations using structured, human-understandable terms. Furthermore, the framework does not require access to the model under investigation, contrarily to
the above techniques which require accessing the layers of a model, circumventing its black-box nature. This white-box access to AI models may be the case in many research scenarios but in the real world, the convenience of tapping into the inner workings of a trained model is unlikely to exist. Black box explanations have been criticised by the community \cite{rudin2019stop}, in favour of inherently interpretable models, however since black box models are still prevalent in both academia and the industry, attempting to explain their predictions is still an important problem to solve. Our proposed solution to the problem mitigates a lot of the criticism, as long as the data and the knowledge graphs are chosen wisely.



Specifically, by considering explanations at the level of abstraction of the knowledge graph, this technique centralizes the possible sources of errors solely toward the data, instead of the explainability algorithm. This transforms typical pitfalls of \textit{post hoc} XAI, such as misleading explanations, to a considerably easier and more intuitive problem. For instance, a pixel-level explanation algorithm, such as a saliency map, explaining why a dog is a labrador and not a golden retriever, would produce a map lighting the body of the dog as an explanation. Due to the ambiguity of such a pixel-level explanation, it may not be apparent which characteristics of the animal's body were vital for its classification - let alone identifying it as an erroneous explanation. In contrast, a semantic explanation would express explicitly that ``the dog's coat has to be red to be classified as a labrador''. The clarity that such an explanation provides is not only more informative but easily exposes possible biases - such as an uneven number of red-coated labradors in the data. Knowing that this bias can only be arising from the data and not the algorithm, makes the resolution straightforward. Namely, choosing 
data and knowledge
which are semantically representative of the classes we want to discern. This is how we can replace the hard problem of algorithmic errors in explainability with the more manageable problem of improper data.

\section{Background and Notation}
  The framework is presented using the formalism of Description Logics (DLs) \cite{DBLP:conf/dlog/2003handbook}. Even though we do not make full use of the expressive power and reasoning capabilities of DLs, they are used as a way of future-proofing the framework, which could be extended in the future for more expressive knowledge than what is presented here. 
  In this work, we make certain assumptions about the structure of DL knowledge bases. Specifically, given a vocabulary $\mathcal{V}=\langle{\mathsf{CN,RN,IN}}\rangle$ where $\mathsf{CN,RN,IN}$ are mutually disjoint finite sets of concept, role and individual names, we consider $\mathcal{K}=\langle{\mathcal{A},\mathcal{T}}\rangle$ to be a knowledge base, where the ABox $\mathcal{A}$ is a set of assertions of the form $C(a)$ and $r(a,b)$ where $C\in\mathsf{CN}$, $r\in\mathsf{RN}$ and $a,b\in{\mathsf{IN}}$, and the TBox $\mathcal{T}$ is a set of terminological axioms of the form $C\sqsubseteq{D}$ where $C,D\in{\mathsf{CN}}$ or $r\sqsubseteq{s}$ where $r,s\in{\mathsf{RN}}$. The symbol `$\sqsubseteq{}$' denotes inclusion or subsumption. For example, a concept name (in $\mathsf{CN}$) could be $\mathsf{Dog}$, an individual name (in $\mathsf{IN}$) could be the (unique) name of a specific dog, for example $\mathsf{snoopy\_42}$, and a role name (in $\mathsf{RN}$) could be a relation, such as ``$\mathsf{eating}$''. Then an ABox could contain the assertion $\mathsf{Dog(snoopy\_42)}$, indicating that $\mathsf{snoopy\_42}$ is a $\mathsf{Dog}$, and a TBox could contain the axiom $\mathsf{Dog}\sqsubseteq{\mathsf{Animal}}$, representing the fact that all dogs are animals (where $\mathsf{Animal}$ is also a concept name in $\mathsf{CN}$).
In such a knowledge base, both the ABox and the TBox can be represented as labeled graphs. 
  An ABox $\mathcal{A}$ can be represented as the graph $\langle V,E,\ell_V,\ell_E\rangle$ (an \emph{ABox graph}),  where $V = \mathsf{IN}$ is the set of nodes, $E = \{ \langle a,b\rangle \mid r(a,b)\in\mathcal{A}\}\subseteq \mathsf{IN}\times \mathsf{IN}$ is the set of labeled edges, $\ell_V: V \rightarrow 2^{\mathsf{CN}}$ with $\ell_V(a) = \{C\mid C(a)\in \mathcal{A}\}$ is the node labeling function, and $\ell_E: E \rightarrow 2^{\mathsf{RN}}$ with $\ell_E(a,b) = \{r\mid r(a,b)\in \mathcal{A}\}$ is the edge labeling function. 
   A TBox $\mathcal{T}$ that only contains hierarchies of concepts and roles, can be represented as a directed graph $\langle{V,E}\rangle$ (a TBox graph) where $V=\mathsf{CN}\cup{\mathsf{RN}}\cup{\{\top\}}$ the set of nodes. The set of edges $E$ contains an edge for each axiom in the TBox, in addition to edges from atoms appearing only on the right side of subsumption axioms, and atoms that don't appear in the TBox, to the $\top$ node: $E=\{\langle{a,b}\rangle{}\mid a\sqsubseteq{b}\in{\mathcal{T}}\}\cup\{\langle{a,\top}\rangle\mid c\sqsubseteq{a}\in{\mathcal{T}}\wedge{a\sqsubseteq{d}}\not\in{\mathcal{T}\wedge{c,d\in{\mathsf{CN}\cup{\mathsf{RN}}}}}\}\cup\{\langle{a,\top}\rangle\mid{a\not\in{sig(\mathcal{T})}}\}$. This is abusive notation, in that the symbol $\top$ is overloaded and symbolizes both the universal concept and the universal role. Finally, we consider classifiers to be functions $F:\mathcal{D}\rightarrow{\mathcal{C}}$, where $\mathcal{D}$ is the domain of the classifier  
   and $\mathcal{C}$ is the set of names of the classes.

\section{Counterfactuals in Terms of Knowledge}
The first step for attempting to understand a black box is to choose what data to feed it. In this work we explore the merits of feeding it data for which there is available information in a knowledge base. This data comes in the form of what we call \textit{exemplars}, that are described as individuals in the underlying knowledge, and can be mapped to the feature domain of the classifier. Such semantic information that describes exemplars can be acquired from knowledge graphs available on the web (for example wordnet \cite{miller1995wordnet}), 
it can be extracted using knowledge extraction methods (such as scene graph generation), or, ideally, it can be provided by domain experts. 
A motivating example would be a set of X-rays that have been thoroughly described by medical professionals, and using standardized medical terminology their characterizations have been encoded in a description logics knowledge base. Having such a set of exemplars allows us to provide explanations in terms of the underlying knowledge instead of being constrained by the features of the classifier. 
\begin{definition}[Explanation Dataset
]  \label{def:explanation dataset}
Let $\mathcal{D}$ be a domain of item feature data, $\mathcal{C}$ a set of classes, and $\mathcal{V}=\langle{\mathsf{IN, CN, RN}}\rangle$ a vocabulary such that $\mathcal{C}\cup\{\mathsf{Exemplar}\} \subseteq \mathsf{CN}$. Let also $\mathsf{EN} \subseteq \mathsf{IN}$ be a set of \emph{exemplars}.
An \emph{explanation dataset} $\mathcal{E}$ in terms of $\mathcal{D}$, $\mathcal{C}$, $\mathcal{V}$ is a tuple $\mathcal{E}= \langle{ \mathcal{M},\mathcal{K}}\rangle$, where $\mathcal{M}:{\mathsf{EN}} \rightarrow \mathcal{D}$ is a mapping from the exemplars to the item feature data, and $\mathcal{K}=\langle\mathcal{T},\mathcal{A}\rangle$ is a DL knowledge base over $\mathcal{V}$ such that $\mathsf{Exemplar}(a)\in \mathcal{A}$ iff $a\in\mathsf{EN}$, the elements of $\mathcal{C}$ do not appear in $\mathcal{K}$, and $\mathsf{Exemplar}$ and the elements of $\mathsf{EN}$ do not appear in~$\mathcal{T}$.
\end{definition}
Intuitively, an explanation dataset contains items for which we have available semantic information, alongside a feature representation that can be fed to the classifier. The concept name $\mathsf{Exemplar}$ is used to flag those individuals that can be mapped by $\mathcal{M}$ to the domain of the classifier, and it does not appear in the TBox to avoid complications that could arise from reasoning.
In this context, counterfactual explanations have the form of \textit{semantic edits} that are applied on an ABox corresponding to an explanation dataset. 
Specifically, given an exemplar and a desired class, we are searching for a set of edits that when applied on the ABox lead to the exemplar being \textit{indistinguishable} from any exemplar that is classified to the desired class. 

\begin{definition}[Counterfactual Explanation]
Let $F:\mathcal{D}\rightarrow{{\mathcal{C}}}$ be a classifier  and $\langle{\mathcal{M},\mathcal{K}}\rangle$ an explanation dataset  where $\mathcal{M}:\mathsf{EN}\rightarrow{\mathcal{D}}$ is a mapping function, $\mathsf{EN}$ is a set of exemplars and $\mathcal{K}=\langle{\mathcal{A},\mathcal{T}}\rangle$ is a knowledge base. A \emph{counterfactual explanation} for an exemplar $a\in{\mathsf{EN}}$ and class $C\in{\mathcal{C}}$ is a tuple $\langle{c,E\rangle}$ where $c\in{\mathsf{EN}}$ and $F(\mathcal{M}(c)) = C$, and $E$ is a set of \emph{edit operations} that when applied on the connected component of $a$ on the ABox graph make it equal to the connected component of $c$. An edit operation on an ABox can be any of:
\begin{itemize}
    \item Replacement of assertion $D(a)$ with $E(a)$, symbolized $e_{D\rightarrow{E}}$
    \item Replacement of $r(a,b)$ with $s(a,b)$, symbolized $e_{r\rightarrow{s}}$
    \item Deletion of $D(a)$ or $r(a,b)$, symbolized $e_{D\rightarrow{\top}}$ or $e_{r\rightarrow{\top}}$ 
    \item Insertion of $D(a)$ or $r(a,b)$, symbolized $e_{\top\rightarrow{D}}$ or $e_{\top\rightarrow{r}}$ 
\end{itemize}
where $D,E\in{\mathsf{CN}}$ and $r,s\in\mathsf{RN}$.
\end{definition}

For example, consider an image classifier $F$ that classifies to the classes $\mathcal{C}=\{\mathsf{WildAnimal,DomesticAnimal}\}$, and two exemplars $e_1,e_2$ each classified to a different class: $F(e_1)=\mathsf{WildAnimal}$ and $F(e_2)=\mathsf{DomesticAnimal}$. The connected components of each exemplar in the ABox graph might be:
\begin{align*}
  \mathcal{A}_{e_1}=\{\mathsf{Exemplar}(e_1),\mathsf{depicts}(e_1,a),\mathsf{depicts}(e_1,b),\\ \mathsf{isIn}(a,b), \mathsf{Animal}(a), \mathsf{Forest}(b)\}  
\end{align*}
\begin{align*}
    \mathcal{A}_{e_2}=\{\mathsf{Exemplar}(e_2),\mathsf{depicts}(e_2,c),\mathsf{depicts}(e_2,d),\\ \mathsf{isIn}(c,d), \mathsf{Animal}(c), \mathsf{Bedroom}(d)\}
\end{align*}
Then an explanation for exemplar $e_1$ and class $\mathsf{DomesticAnimal}$ would be the replacement of assertion $\mathsf{Forest}(b)$ with $\mathsf{Bedroom}(b)$, which would be symbolized $\langle{e_2,\{e_{\mathsf{Forest}\rightarrow\mathsf{Bedroom}}\}}\rangle$ and it would be interpreted by a user as ``If image $e_1$ depicted animal $a$ in a $\mathsf{Bedroom}$ instead of a $\mathsf{Forest}$, then the image would be classified as a $\mathsf{DomesticAnimal}$''. 
Of course there is no way to know if the image $e_1$ with the $\mathsf{Forest}$ replaced with a $\mathsf{Bedroom}$ would be classified to the target class, because we do not have a way to edit the pixels of the image and feed it to the classifier. The explanation however provides useful information to the user and can potentially aid in the detection of biases of the classifier. For example, after viewing this explanation, the user might choose to feed the classifier images depicting wild animals in bedrooms to see whether or not they are misclassified as domestic animals.

To provide more information to the end user, we can accumulate counterfactual explanations for multiple exemplars and the desired class and provide statistics about what changes tend to flip the prediction of the classifier, as a form of a ``global'' explanation. For example, one could ask ``What are the most common semantic edits that when applied on exemplars depicting bedrooms lead to them to be classified as wild animals?''. To do this, we first compute the multiset $\mathcal{G}$ of all counterfactual explanations from each exemplar in the source subset 
to the target class
, and then we show the end-user the \textit{importance} of each atom for changing the prediction on the source exemplars to the target class, where
$$\mathsf{Importance}(y)=\frac{|\{e_{x\rightarrow{y}}\in{\mathcal{G}}\}|-|\{e_{y\rightarrow{x}}\in{\mathcal{G}}\}|}{|\mathcal{G}|}$$
where $, x,y\in{\mathsf{CN}}\text{, or }x,y\in{\mathsf{RN}}$.

Intuitively, the importance of an atom shows how often it is introduced (either via replacement or via insertion) as part of the semantic edits of a set of counterfactual explanations. A negative importance would indicate that the atom tends to be removed (either via replacement or via deletion of assertions). For example, one could gather all exemplars that are classified as $\mathsf{WildAnimal}$, along with their counterfactual explanations for target class $\mathsf{DomesticAnimal}$ and compute how important the presence (or absence) of a concept or a role is for distinguishing between the two classes.

\section{Computing Counterfactual Explanations}
\label{sec:algo}
Given an explanation dataset $\langle{\mathsf{EN}\rightarrow{\mathcal{D}},\langle{\mathcal{A},\mathcal{T}}\rangle}\rangle$, the first step for computing counterfactual explanations is to determine the edit operations on the ABox that transform the description of every exemplar to every other exemplar, thus this is a computation that has to be done $O(|\mathsf{EN}|^2)$ times, but it only has to be done once for an explanation dataset. Ideally, each set of edit operations will be minimal as they are intended to be shown to users as explanations, which means that the problem to be solved is the exact graph edit distance problem \cite{GED}.
\subsection{Edit Distance Between Exemplars}
 Unfortunately, computing the graph edit distance is NP-hard \cite{zeng2009comparing}, and even though there are optimized algorithms for its computation \cite{abu2015exact}, it will not be feasible for explanation datasets with a large number of exemplars. 
One way to overcome the complexity is to simplify the problem, and to work with \emph{sets} instead of graphs, which will allow us to use an algorithm similar to the one presented in \cite{filandrianos2022conceptual} for the computation of explanations. Of course converting a graph into a set without losing information is not generally possible. In this work, we convert the connected components of exemplars on the ABox graph into \textbf{sets of sets} of concepts, by rolling up the roles into concepts. Specifically, we add information about \textit{outgoing edges} to the label of each node in the ABox graph, by defining new concepts $\exists{r.C}$ for each pair of role name $r$ and concept name $C$, and then adding $\exists{r.C}$ to the label of a node $a$ if $r(a,b),C(b)\in{\mathcal{A}}$ for any $b\in{\mathsf{IN}}$. Then every exemplar of the explanation dataset is represented as the set of labels of nodes that are part of the connected component of the exemplar on the ABox. For instance, an exemplar $e$ with a connected component: $\mathcal{A}_e=\{\mathsf{Exemplar}(e),\mathsf{depicts}(e,a),\mathsf{depicts}(e,b),\mathsf{depicts}(e,c),\mathsf{Cat}\\(a),\mathsf{eating}(a,b),\mathsf{Fish}(b),\mathsf{in}(b,c)$, $\mathsf{Water}(c)$
would be represented as the set of labels (ignoring the Exemplar node):
$\{\{\mathsf{Cat},\exists{\mathsf{eating}.\mathsf{Fish}}\},$
$
    \{\mathsf{Fish},\exists{\mathsf{in}}.\mathsf{Water}\}, \{\mathsf{Water}\}\}$. Now, to compute counterfactual explanations, we have to solve a \textit{set edit distance} problem.
\subsection{Cost of Edits}
\label{sec:algo-cost}
Before solving the edit distance problem we first have to determine how much each edit costs. Intuitively, we want counterfactual explanations to be semantically similar exemplars, thus the cost of an edit should reflect how much the exemplar changes semantically after applying the edit. 
To do this, we utilize the information that is present in the TBox. 
For the first type of ABox edits, that involves replacing concept assertions $(e_{A\rightarrow{B}})$, we assign a cost to the replacement of concept $A$ with concept $B$ equal to their distance on the TBox graph, ignoring the direction of the edges. For example, given a TBox:
$\mathcal{T}=\{\mathsf{Cat}\sqsubseteq{\mathsf{Mammal}}, \mathsf{Dog\sqsubseteq{\mathsf{Mammal}}},\mathsf{Ant}\sqsubseteq{\mathsf{Insect}}, \mathsf{Mammal}\sqsubseteq{\mathsf{Animal}}, \mathsf{Insect}\sqsubseteq{\mathsf{Animal}}\}$
the cost of replacing a $\mathsf{Cat}(a)$ assertion with $\mathsf{Mammal}(a)$ would be $1$, the cost of replacing $\mathsf{Cat}(a)$ with $\mathsf{Dog}(a)$ would be $2$ and the cost of replacing $\mathsf{Cat}(a)$ with $\mathsf{Ant}(a)$ would be $4$. Similarly, the cost of replacing a role assertion $r(a,b)$ with $s(a,b)$ (symbolized $e_{r\rightarrow{s}}$) is assigned to be the distance of the shortest path on the undirected TBox graph from $r$ to $s$. It is worth mentioning that this is not necessarily the optimal way to compute semantic similarities of concepts and roles, and other measures exist in the literature \cite{d2009semantic}. 
For the insertion of concept or role assertions, as is apparent from the notation $e_{\top\rightarrow{a}}$, we assign a cost equal to the distance of the inserted atom (either a role or a concept) from the $\top$ node in the TBox graph. This means that it is more expensive to insert more specific atoms, than more general ones. Similarly for the deletion of atoms $e_{a\rightarrow\top}$, the cost is assigned to be the distance of the deleted concept or role from the $\top$ node on the undirected TBox graph. 
Finally, we allow a user to manually assign cost to edits which could be useful in specific applications where some edits might not be feasible in the real world. For example, if we had exemplars representing people, and concepts representing their age ($\mathsf{Young},\mathsf{Old}$) we might want to disallow the edit $e_{\mathsf{Old\rightarrow{Young}}}$ as it would require time-travel in order to be implemented realistically, so we could assign an infinite cost to this edit.

\subsection{Algorithm}

In the general case, the algorithm for computing counterfactual explanations has two steps. The first step (preprocessing) is to compute the edit path between all pairs of exemplars in an explanation dataset and to acquire predictions of the classifier on all exemplars. 
The second step is, given an exemplar and a target class, to find the exemplar with the minimal edit distance that is classified to the target class. 
For the case of graph edits, in our experiments we use a depth-first graph edit distance algorithm proposed in \cite{abu2015exact}, as it is implemented in the networkx python package\footnote{\url{https://networkx.org/documentation/stable/reference/algorithms/generated/networkx.algorithms.similarity.optimize_graph_edit_distance.html}}.
For the case of \textit{concept set descriptions}, first we need to find the connected components of exemplars on the ABox graph. Then we need to add $\exists{r.C}$ concepts to the labels of nodes $a$ for which $r(a,b)C(b)$ is in the ABox. 
To compute the set edit distance between two labels of nodes $\ell{}_a,\ell{}_b$, each of which is a set of concepts (either atomic or of the form $\exists{r.C}$), we first construct a bipartite graph where each element of $\ell{}_a$ is connected to every element of $\ell{}_b$ and has a cost based on the TBox $\mathcal{T}$, as defined in section \ref{sec:algo-cost}. On this bipartite graph we then compute the minimum weight full match using an implementation of Karp's algorithm \cite{karp1980algorithm} for the problem  to get the optimal set of edits from one set of concepts to another. 
Finally, to compute the edit distance between two sets of labels $L_1,L_2$, each of which is \textbf{a set of sets} of concepts, we first compute the edit distance from each label in $L_1$ to every label in $L_2$ by using the procedure described in the previous paragraph for each pair of labels, meaning the set edit distance computation is performed $|L_1||L_2|$ times. Then to find the edit distance between $L_1$ and $L_2$ we use the same procedure as with sets of concepts (bipartite graph and full match), but this time the weights of the edges of the bipartite graph are assigned according to set the edit distance. Having preprocessed the explanation dataset and saved the edit paths, an explanation can be provided in $O(\mathsf{|EN|})$. Regarding the complexity of the preprocessing algorithm, we refer to the supplementary material \footnote{https://github.com/geofila/Semantic-Counterfactuals/blob/main/Supplementary\%20Material.pdf}.

\section{Experiments}
For evaluating the proposed framework, we conducted four experiments, each with a different purpose. The first is a user study for comparing our work with a state-of-the-art image counterfactual system, which was performed on the CUB dataset \cite{wah2011caltech}. The second is a demonstration of an intended use-case of the framework, where in order to explain a black-box classifier trained on the Places dataset \cite{places2018}, we utilize semantic information present in COCO \cite{lin2014microsoft}, the Visual Genome \cite{krishna2017visual}, and WordNet. In the third experiment we explore the use of a scene graph generator for producing semantic descriptions. For the final experiment, we generate explanations for a COVID-19 cough audio classifier trained on a subset of Coswara \cite{sharma2020coswara}, showcasing that our approach is domain agnostic, and how it can provide useful explanations in such a critical application. 



\subsection{Human Evaluation on the CUB Dataset}
To assess how the counterfactual images retrieved by our algorithm fare against the state-of-the-art results \cite{goyal2019counterfactual}, we set up a human study; since a widely accepted metric to evaluate the success of semantically consistent visual counterfactuals does not exist.

\subsubsection{Setting}
We first acquire two pre-trained classifiers (a VGG-16 \cite{simonyan2013deep}, and a ResNet-50 \cite{he2016deep}), and make predictions on the test set of CUB. This dataset is what we use as an \textit{explanation dataset}, after encoding the annoatations of the images in a DL knowledge base. 
 
 In \cite{vandenhende2022making}, the authors selected a number of bird images from the CUB dataset. Then, for each one, they retrieved its closest counterfactual image from the full dataset, with the restriction that it cannot belong to the same bird species (label) as the source. For our experiment, we executed the same methodology utilizing our algorithm to perform the same task on the same source images. 
 
Then, to each of our 33 human evaluators, we presented a randomly selected source image along with its two corresponding counterfactual images - the one retrieved by the SOTA and by our algorithm. The evaluators were then asked which of the two counterfactual bird images more closely, semantically, resembled the bird depicted in the first image (i.e. not taking into account the bird's posture or its background).

\subsubsection{Results}
The images retrieved with both methods were largely similar and sometimes identical. As a result, the evaluators experienced difficulties deciding between the two counterfactual images, and the two methods achieved similar results (Table \ref{table:human}). It is important to note that our algorithm did not peek inside the model, contrarily to the SOTA algorithm. Our approach managed to attain equal results just by taking into account the semantic knowledge accompanying CUB images, without having white-box access to the classifiers. Further details about this experiment are available in the supplementary material$^2$.

\begin{table}[]
\centering
\resizebox{0.48\textwidth}{!}{
\begin{tabular}{l|l|l}
                 & ResNet-50 & VGG-16  \\ \hline \hline
\cite{vandenhende2022making} S.O.T.A. & 14.65\%   & 13.68\% \\
Ours              & 34.93\%   & 23.65\% \\
Can't Tell        & \textbf{50.42\%}   & \textbf{62.67\%}

\end{tabular}
}
\caption{Human evaluation results on which of the two counterfactual bird images is semantically closer to the source image.}
\label{table:human}
\end{table}

\begin{figure}[]
    \centering
    \includegraphics[scale=0.2]{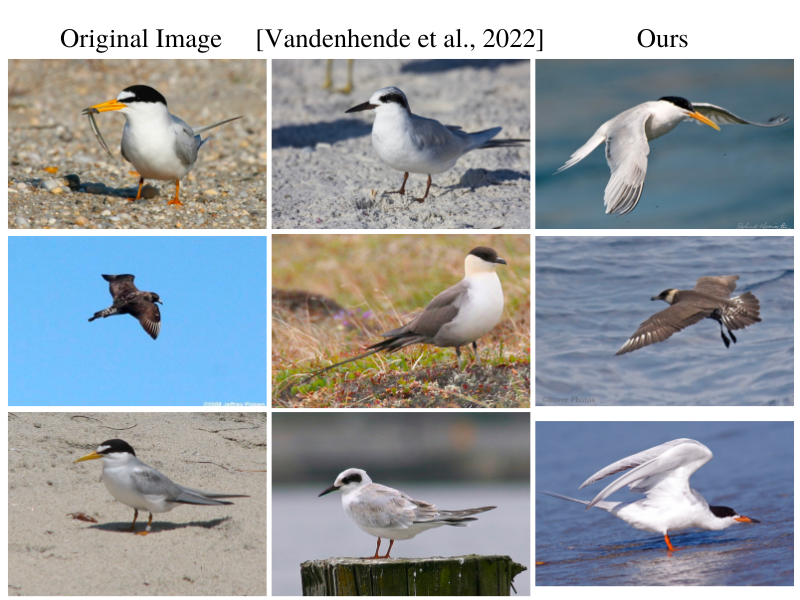}
    \includegraphics[scale=0.2]{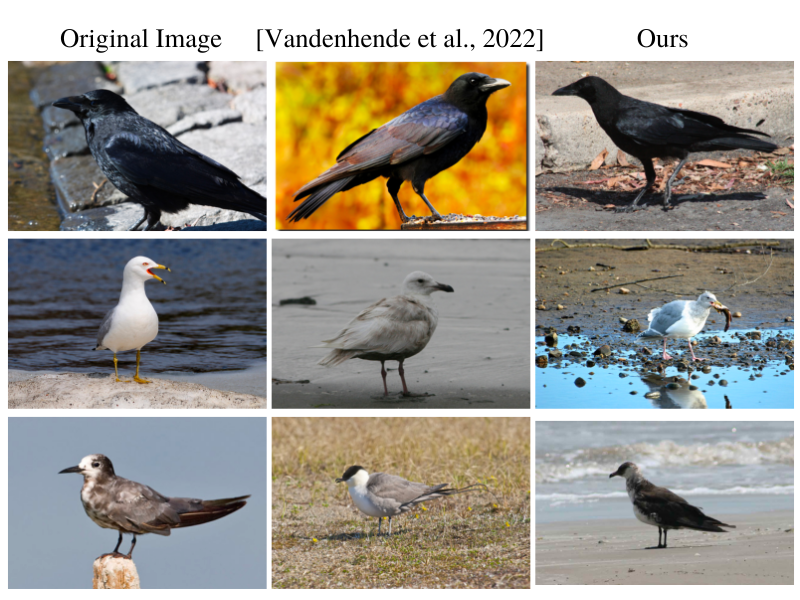}
    \caption{The first column shows the original image, the second one \protect\cite{vandenhende2022making}'s retrieved image and the third one the image retrieved by our algorithm.}
    \label{fig:human-examples}
\end{figure}

\subsection{Explaining a Places Classifier Using COCO}
For our second experiment, we decided to explore more realistic examples and took advantage of the COCO dataset, which contains object-annotated, real-world images that can automatically be linked to an external knowledge.

\subsubsection{Setting}
We initially examined COCO's labels to determine which class transitions we could utilize for our counterfactuals, and concluded that the two classes should be ``Restaurant'' related and ``Bedroom'' related images. Details about the subset of COCO can be found in the supplementary material $^2$.
For each image, a description of the objects present in that image is provided. To create the explanation dataset, we automatically linked these object descriptions with WordNet synsets by using the NLTK python package\footnote{https://www.nltk.org/howto/wordnet.html}. We used WordNet synsets as the set of concept names $\mathsf{CN}$, and the hyponym-hypernym hierarchy as a TBox. 
We then selected an image classifier, trained specifically for scene classification on the PLACES dataset, provided by its creators \footnote{http://places2.csail.mit.edu/index.htm}, and made predictions on the aforementioned subset of COCO. This is the black-box classifier which we provide explanations for.

\subsubsection{Results}
\begin{figure}[]

\centering
\scalebox{0.467}{
\minipage{1\textwidth}
  \includegraphics[width=\linewidth]{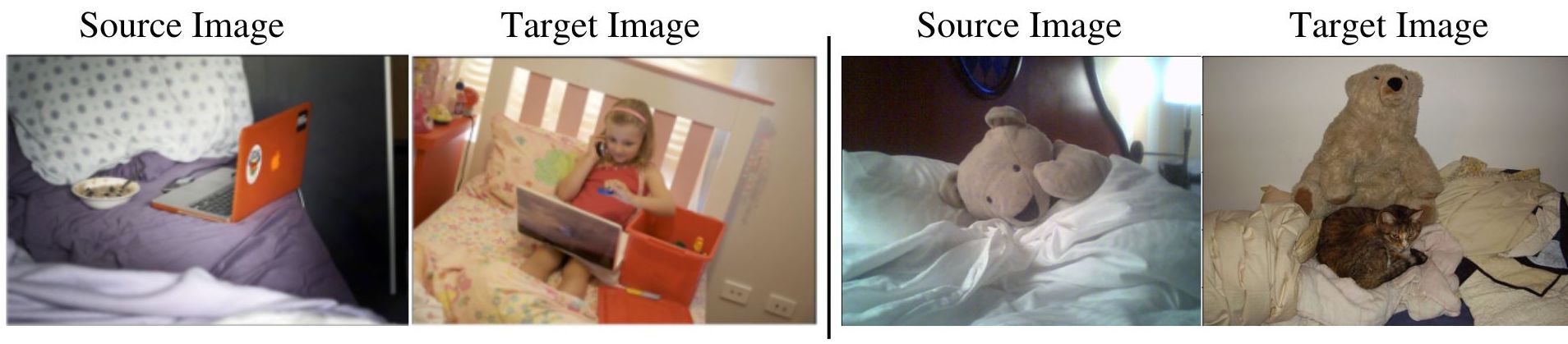}
\endminipage\hfill
}
\caption{Counterfactual explanations for changing the predictions of for two images. For `Bedroom' to `Playhouse' is to simply to add a child ($e_{\top\rightarrow{\mathsf{Child}}}$) (left) and for `Bedroom' to `Veterinarians Office' is to add a cat ($e_{\top\rightarrow{\mathsf{Cat}}}$) (right).}
\label{fig:playhouseVet}
\end{figure}

In the first row of fig.\ref{fig:playhouseVet} we show a local counterfactual explanation for an image classified as a ``Bedroom'' to the target class ``Playhouse'', which requires only one Concept Edit ($e_{\top\rightarrow{\mathsf{Child}}}$). This example is interesting because ``Playhouse'' is an erroneous prediction (the ground truth for the second image should be ``Bedroom''), thus immediately we detect a potential bias of the classifier, that if a $\mathsf{Child}$ is added to an image of a ``Bedroom'' it might be classified as a ``Playhouse''. Similarly, in the second row of fig.\ref{fig:playhouseVet} we show a counterfactual for an image classified as ``Bedroom'' to the target class ``Veterinarian's Office'', and the resulting target image is again an erroneous prediction. 

\widowpenalty=0
\clubpenalty=0

\begin{figure}[]
    \centering
    \includegraphics[scale=0.5]{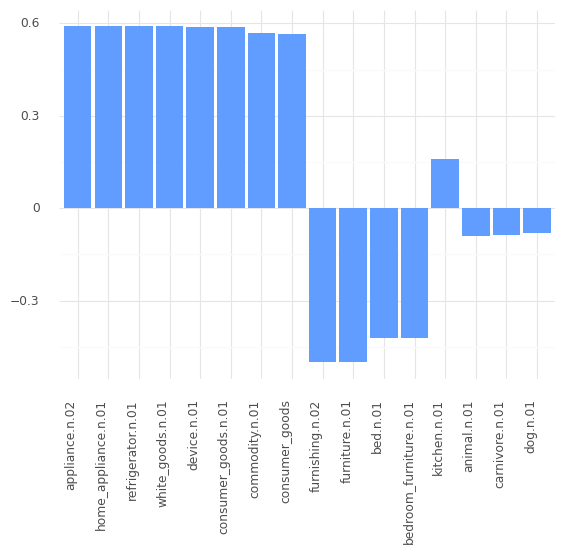}
    \caption{Global explanation for  the subset of COCO which is classified as ``bedroom'', with target class ``kitchen''}
    \label{fig:global_coco_a}
\end{figure}

\begin{figure}[]
    \centering
    \includegraphics[scale=0.5]{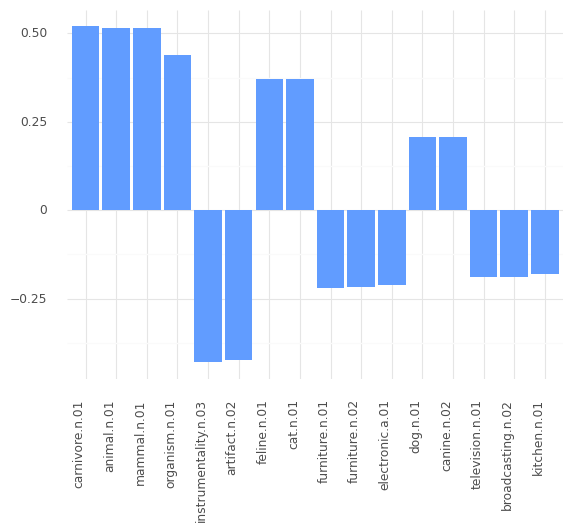}
    \caption{Global explanation for the subset of COCO which is classified as ``bedroom'', with target class ``veterinarian''}
    \label{fig:global_coco_b}
\end{figure}

In fig.\ref{fig:global_coco_a} and fig.\ref{fig:global_coco_b} we see two examples of global counterfactual explanations 
on the COCO dataset. 
As an unbiased exercise, we can try to work out the source and target classes for each figure, just by looking at the most frequent additions and removals it contains. On the first (fig.\ref{fig:global_coco_a}), which is the simpler of the two, we see that the most common removals from the source images were concepts relevant to \{furniture, bed, animal, carnivore, dog\}, while the most common additions were the concepts \{home appliance, refrigerator, white goods, consumer goods\}. From this, we can assume that the source subset more likely contained bedroom images (with a bias towards pets) and the target class was probably a kitchen. The true classes were, indeed, ``bedroom'' and ``kitchen''.
On the second (fig.\ref{fig:global_coco_b}), we see that the most frequent removals revolved around furniture and electronics, 
and the most common additions around animals. 
Knowing that we are dealing with a classifier of rooms and places, we would probably guess a kitchen for the source and some type of location with a lot of domestic animals for the target. The actual classes were ``bedroom'' (surprisingly) and ``veterinarian office'', which raises the interesting question: why did we see removals of ``kitchens'' instead of ``beds'' from the bedroom class? And the answer is: because no beds were really removed since veterinarian office images tend to include beds. Moreover, trying to understand the ``kitchen'' removal from bedroom photos and browsing through training dataset, we notice that it contains several studio-apartment bedroom images that have part of their kitchen appearing in the photo - kitchens that are mostly missing from vets' offices and thus had to be removed.

\subsubsection{Comparison With Visual Genome’s Results}
\begin{figure}[]
    \centering
    \includegraphics[scale=0.5]{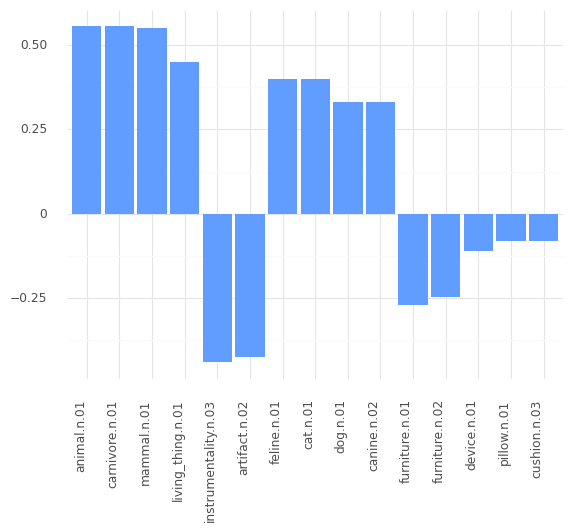}
    \caption{Global explanation for the subset of Visual Genome which is classified as ``bedroom'', with target class ``vet''}
    \label{fig:global_visual_b}
\end{figure}

A crucial question at this point is how can we know where the biases that our system uncovers come from. We are assuming that they emerge from the classifier, but a biased explanation dataset could yield similarly biased results. A way to answer this question is to run the same task on a different dataset, to see how those results compare with the previous ones. 
For our cross-checking dataset, we will use Visual Genome since it is, along with COCO, one of the very few datasets containing annotated images. The results of the Visual Genome experiment, overlayed on COCO's results, are depicted in fig. \ref{fig:global_visual_b}. We can see that the classifier gave very similar predictions for both datasets
, which validates the hypothesis that the biases did not arise from a possible irregular distribution within the explanation datasets but from the classifier itself.

\subsection{Testing the Importance of Roles}
Most of the important features that differentiated the classes in the previous experiment could be fully expressed by concepts alone, e.g. the existence of a bed or a dog. There are, though, many situations where this is not the case and where roles and relationships between objects should be taken into account. For example, classifying between ``driver'' and ``pedestrian'' classes on images containing the concepts ``motorbike'', ``bicycle'' and ``person'' cannot be done without knowing the relationship between the person and the vehicle. 

\subsubsection{Setting}
The issue with this experiment was the general unavailability of datasets that include images along with their semantic descriptions, which is critical information for our system to work. Visual Genome does include roles, but they are few and inconsistent since they may be present in one photo but missing in other similar ones. Moreover, most real-world applications would not have knowledge accompanying their image datasets. 
To tackle this practical obstacle, we decided to use a Scene Graph Generator \cite{SGG2022} that can extract concepts and roles from images. 
Details about the parameters used for scene graph generation are in the supplementary material $^2$.


The first step is to search the web for images satisfying our criteria and divide them into two classes, namely ``driver'' and ``pedestrian''. We do this for motorbike and bicycle riders since we want to avoid the role name being itself the descriptor of the class, e.g. ``person driving car''. We query Google, Bing and Yahoo images for a combination of keywords containing ``people'', ``motorbikes'' and ``bicycles'', gather the following creative-commons photographs, and manually split them into two classes.
1. $\{\text{driver class}\}$ (63 images of people on bicycles and 127 images of people on motorbikes)
2. $\{\text{pedestrian class}\}$ (31 images of people and parked motorbikes, 38 images of people and parked bicycles). 
Once we construct our dataset, we  extract semantic descriptions with the scene graph generator. 


\subsubsection{Results}
\begin{figure}[]
    \centering
    \includegraphics[width=0.9\linewidth]{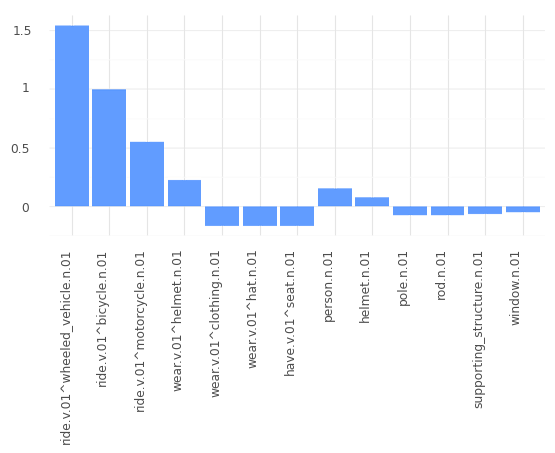}
    \caption{Flipping class form ``pedestrian'' to ``driver'', the most important changes are: the addition of ``ride\string^wheeled\string_vehicle'', ``wear\string^helmet'' and the removal of ``wear\string^hat''.}
    \label{fig:driving_edges}
\end{figure}

The global counterfactuals transitioning from ``pedestrian'' to "driver", are depicted on fig.\ref{fig:driving_edges} as concept set descriptions, i.e. concepts along with roles. The top addition by a very large margin is ``$ride\string^wheeled\string_vehicle$'' as expected, which is the parent, and thus, the sum of ``$ride\string^bicycle$'' and ``$ride\string^motorbike$''. Next, we see additions of ``$wearing\string^helmet$'' and a smaller addition of the concept ``helmet'' by itself, presumably because in some driving photos the helmet was on the handlebars of the bike and not on the rider's head. We also see that ``$wear\string^hat$'' is removed (the child of ``$wear\string^clothing$''), which compliments the addition of ``$wear\string^helmet$'', and that ``$have\string^seat$'' is removed since bicycle seats are not visible when bikes are ridden. The rest of the edits are too scarce and, although we might be able to explain them, they can very likely be noise as well.

\subsection{COVID-19 Classification}
Our final experiment showcases the framework in the audio domain, and specifically in an application where explainability is crucial: COVID-19 diagnosis.

\subsubsection{Setting}
We provide explanations for a classifier that was trained on a subset of the Coswara Dataset, specifically, the winning entry of the IEEE COVID-19 sensor informatics challenge \footnote{https://healthcaresummit.ieee.org/data-hackathon/ieee-covid-19-sensor-informatics-challenge/}. The input of the classifier is an audio file of a person's coughing, and the output is the probability that the user to has the COVID-19 virus. As an \textit{explanation dataset}, we used data from the Smarty4covid platform \footnote{https://doi.org/10.5281/zenodo.7137424}, which contains similar audio files and includes additional annotations, such as gender, symptoms, medical history, etc., in the form of an ontology. It is worth mentioning that as features of the audio files, we selected only the concepts that can be expressed in the audio file. Thus, we removed concepts such as vaccination status.

\subsubsection{Results}
The global counterfactuals transitioning from ``COVID-19 Negative'' to ``COVID-19 Positive'' (Table \ref{table:covid}) depicts that the top insertion is the concept ``Symptom'', which is the parent of all the symptoms of the knowledge base. However, not every symptom is capable of altering the prediction of the classifier since the concept ``Respiratory'' which is a child of the concept ``Symptom'' and the parent of all the symptoms that are related to the respiratory system (e.g., ``Dry Cough'') is the next most added concept along with its children such as ``Dry Cough'', ``Runny Nose'', and ``Cough''. In this experiment, we also uncovered an unwanted bias of the classifier since one of the most common edits was to change the user's sex. After this peculiar observation, we conducted a search on the training dataset, and we found out that this bias was inherited from the training set of the classifier. In particular, on the Coswara dataset, 42\% of females are COVID-19 positive, while for males the percentage is 27\%, which made the classifier erroneously correlate sex to COVID-19 status. 

\begin{table}[]
\begin{tabular}{cc|cc}
Concept     & Importance & \multicolumn{1}{l}{Concept} & \multicolumn{1}{l}{Importance} \\ \hline \hline
Symptom     & -1.298     & Runny Nose                  & -0.22                          \\
Respiratory & -1.278     & Dry Cough                   & -0.19                          \\
Female      & 0.25       & Cough                       & -0.189                         \\
Male        & -0.254     & Sore Throat                 & -0.13                         
\end{tabular}
\caption{The global counterfactuals transitioning from ``COVID-19 Negative'' to ``COVID-19 Positive'', for a classifier trained on coughing audios from Coswara Dataset, using Smarty4covid data as the Explanation Dataset.}
\label{table:covid}
\end{table}

\section{Conclusion}
We have presented a novel explainability framework based on knowledge graphs. The explanations are guaranteed to be valid, and feasible, as they are always edits towards real data points. They are also guaranteed to be minimal, as they are the result of an edit distance computation, and actionable, provided the manual assignment of edit costs. Via the human study, we also show that counterfactual explanations in the context of the proposed framework, are understandable, and satisfactory to end-users. 
The main limitation of the framework, is its dependence on the explanation dataset, which ideally is curated by domain experts. For critical applications, such as medicine, we argue that it is worth the resources. For other applications, we have shown that using available semantically enriched datasets, such as the visual genome, or using automatic knowledge extraction techniques to construct the explanation dataset, such as scene graph generation, can lead to useful explanations.
There are two directions we plan to further explore in future work. Firstly, we aim to extend the framework for more expressive knowledge, and to make use of theoretical results regarding description logics and reasoning. Secondly, we are experimenting with generative models, that can apply the semantic edits on a data sample, and generate a new sample that can be fed to the classifier.

\appendix

\section*{Contribution Statement}
The work presented in this paper is an equal contribution by Edmund Dervakos, Konstantinos Thomas, and Giorgos Filandrianos.


\bibliographystyle{named}
\bibliography{ijcai23}

\end{document}